\documentclass[10pt,twocolumn,letterpaper]{article}

\usepackage[pagenumbers]{cvpr} 
\usepackage{multirow}
\usepackage{makecell}
\usepackage[dvipsnames]{xcolor}

\usepackage{bbding}
\usepackage{bbm}
\usepackage{amssymb}
\usepackage{colortbl}
\definecolor{cvprblue}{rgb}{0.21,0.49,0.74}
\usepackage[pagebackref,breaklinks,colorlinks,citecolor=cvprblue]{hyperref}

\def\paperID{*****} 
\def\confName{CVPR}
\def\confYear{2024}

\title{ Pedestrian Attribute Recognition: A New Benchmark Dataset and A Large Language Model Augmented Framework }

\author{Jiandong Jin${^1}$, Xiao Wang*${^2}$, Qian Zhu${^2}$, Haiyang Wang${^2}$, 
        Chenglong Li${^1}$\thanks{Corresponding Author: Xiao Wang, Chenglong Li} \\ 
${^1}$ {School of Artificial Intelligence, Anhui University, Hefei, China} \\ 
${^2}$ {School of Computer Science and Technology, Anhui University, Hefei, China} \\
\textit{\{jdjinahu, wangxiaocvpr, lcl1314\}@foxmail.com, \{zq542664, why2434961256\}@163.com} 
}

\begin{document}
\maketitle

\begin{abstract}
Pedestrian Attribute Recognition (PAR) is one of the indispensable tasks in human-centered research. However, existing datasets neglect different domains (e.g., environments, times, populations, and data sources), only conducting simple random splits, and the performance of these datasets has already approached saturation. In the past five years, no large-scale dataset has been opened to the public. To address this issue, this paper proposes a new large-scale, cross-domain pedestrian attribute recognition dataset to fill the data gap, termed MSP60K. It consists of 60,122 images and 57 attribute annotations across eight scenarios. Synthetic degradation is also conducted to further narrow the gap between the dataset and real-world challenging scenarios. To establish a more rigorous benchmark, we evaluate 17 representative PAR models under both random and cross-domain split protocols on our dataset. Additionally, we propose an innovative Large Language Model (LLM) augmented PAR framework, named LLM-PAR. This framework processes pedestrian images through a Vision Transformer (ViT) backbone to extract features and introduces a multi-embedding query Transformer to learn partial-aware features for attribute classification. Significantly, we enhance this framework with LLM for ensemble learning and visual feature augmentation. Comprehensive experiments across multiple PAR benchmark datasets have thoroughly validated the efficacy of our proposed framework. 
The dataset and source code accompanying this paper will be made publicly available at \url{https://github.com/Event-AHU/OpenPAR}. 
\end{abstract}

\section{Introduction} 
Pedestrian Attribute Recognition (PAR)~\cite{wang2022PARsurvey} has been widely exploited in the Computer Vision (CV) and Artificial Intelligence (AI) community. It aims to map the given pedestrian image into semantic labels, such as \textit{gender}, \textit{hairstyle}, and \textit{wearings}, using deep neural networks and achieves high performance on current benchmark datasets. These models can be employed in practical scenarios and may work well in simple scenarios. It can also help other human-centric tasks, e.g., pedestrian detection and tracking~\cite{li2024attmot}, person re-identification~\cite{lin2019improving} and retrieval~\cite{huang2024attPersonRetrieval}. 
However, the performance of the current PAR model is still significantly affected by challenging factors (e.g., low illumination, motion blur, and complex backgrounds); moreover, there is still much room for exploration in the relationship between pedestrian image perception and multi-label attributes.

\begin{figure*}
\centering
\includegraphics[width=\textwidth]{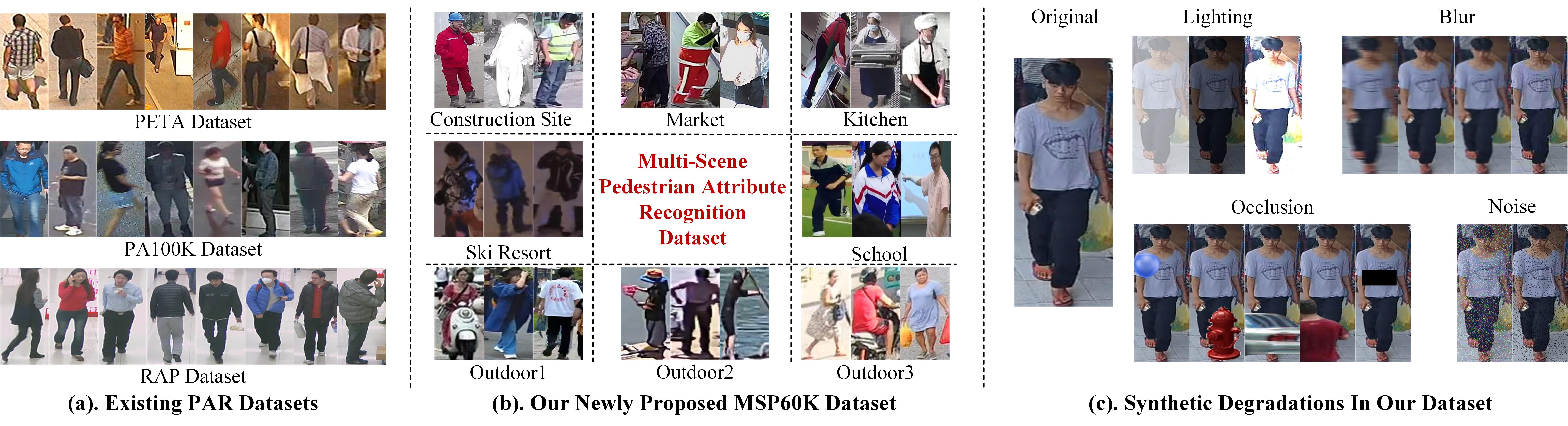}
\caption{(a, b). Comparison between existing PAR datasets and our newly proposed MSP60K dataset. (c). Illustrates the synthetic degradation challenges we employed in our dataset to simulate the complex and dynamic real-world environment.} 
\label{firstIMG}
\end{figure*} 


Considering these issues, we meticulously review the existing works and datasets on PAR and find that the development in the PAR field has begun to enter a bottleneck period. As an effective driving force for promoting the development of PAR, benchmark datasets play a crucial role. However, we believe that the PAR community needs to address several core issues on the benchmark datasets as follows: 
\textbf{1).} The performance of existing pedestrian attribute recognition datasets is \textit{close to saturation}, and the performance improvement of new algorithms has shown a trend of weakening. However, only one small-scale PAR-related dataset has been released in the past five years, thus, there is an urgent need for new large-scale datasets to support new research endeavors.   
\textbf{2).} Existing PAR datasets use random partitioning for model training and testing, which can measure the overall recognition capability of a PAR model. However, this partitioning mechanism overlooks the impact of \textit{cross-domain} (e.g., different environments, times, populations, and data sources) on the PAR model. 
\textbf{3).} Existing PAR datasets do not prominently reflect challenge factors, thus, this may potentially result in neglecting the impact of \textit{data corruption} during real-world application, thereby introducing safety hazards in practical settings. 
In conclusion, it is evident that the PAR community urgently requires a new large-scale dataset to bridge the existing data gap.

In this paper, we propose a new benchmark dataset for pedestrian attribute recognition, termed \textbf{MSP60K}, as shown in Fig.~\ref{firstIMG}. It contains 60,122 images, and over 5,000 person IDs, collected using smart surveillance systems and mobile phones. To make our dataset better reflect the challenges found in real-world scenarios, in addition to annotating as many complex images as possible, we also process these images using additional destructive operations, including blur, occlusion, illumination, adding noise, jpeg compression, etc. As these images belong to different domains and scenarios, such as supermarket, kitchen, construction site, ski resort, and various outdoor scenes, we split these images according to two protocols, i.e., \textit{random split} and \textit{cross-domain split}. Therefore, the newly proposed benchmark dataset can better validate the performance of PAR models in real-world scenarios, especially under cross-domain settings. To build a more comprehensive benchmark dataset for pedestrian attribute recognition, we also train and report 17 representative and recently released PAR algorithms. These benchmark comparison methods can better facilitate the subsequent verification and experimentation of future PAR models.

Based on our newly proposed MSP60K PAR dataset, we also propose a novel large language model (LLM) augmented pedestrian attribute recognition framework, termed LLM-PAR. 
Based on the widely used multi-label classification framework, we rethink the relationship between pedestrian image perception and large language models as the key insight of this work. As we all know, large language models possess powerful abilities in text generation, comprehension, and reasoning. Therefore, we introduce a large language model, which generates textual descriptions of the image's attributes as an auxiliary task based on a multi-label classification framework. This LLM branch serves a dual purpose: on the one hand, it can assist in the learning of visual features through the generation of accurate textual descriptions, thereby achieving high-performance attribute recognition; on the other hand, the LLM can facilitate effective interaction between visual features and prompts. The output text tokens can also be integrated with the aforementioned multi-label classification framework for ensemble learning. 


As shown in Fig.~\ref{framework}, our proposed LLM-PAR can be divided into two main modules, i.e., the standard multi-label classification branch and the large language model augmentation branch. Specifically, we first partition the given pedestrian image into patches and project them into visual embeddings. Then, a visual encoder with LoRA~\cite{hu2022lora} is utilized for global feature learning and a \textbf{M}ulti-\textbf{E}mbedding \textbf{Q}uery Trans\textbf{Former} (MEQ-Former) is proposed for part-aware feature learning. After that, we adopt CBAM~\cite{woo2018cbam} attention modules to merge the output tokens and feed them into MLP (Multi-Layer Perceptron) layers for attribute classification. More importantly, we concatenate the part-aware visual tokens with the instruction prompt and feed them into the large language model for pedestrian attribute description. The text tokens are also fed into an attribute recognition head and ensembles with classification logits. Extensive experiments on our newly proposed MSP60K dataset and other widely used PAR benchmark datasets all validated the effectiveness of our proposed LLM-PAR.

To sum up, we draw the main contributions of this paper as the following three aspects: 

1). We propose a new benchmark dataset for pedestrian attribute recognition, termed MSP60K, which contains 60122 images, over 5,000 IDs, and fully reflects the key challenges in real-world scenarios. We benchmark 17 PAR algorithms on the MSP60K dataset and hope that the introduction of this benchmark dataset can better promote the development and practical deployment of PAR models.

2). We propose a novel large language model (LLM) augmented PAR algorithm, termed LLM-PAR, based on the standard multi-label classification framework. The introduction of the LLM branch enables PAR to better leverage its reasoning capabilities, achieving enhanced visual feature representation and model integration.

3). Extensive experiments conducted on our newly proposed MSP60K dataset and other PAR datasets fully demonstrate the effectiveness of our proposed PAR model. New state-of-the-art performances are achieved on multiple PAR datasets, e.g., 92.20/90.02 on mA/F1 metric on the PETA dataset, 91.09/90.41 on PA100K.

\section{Related Works} 

\subsection{Pedestrian Attribute Recognition} 
Pedestrian attribute recognition~\cite{wang2022PARsurvey}\footnote{\url{https://github.com/wangxiao5791509/Pedestrian-Attribute-Recognition-Paper-List}} aims to classify pedestrian images based on a predefined set of attributes. Current methods can be broadly categorized into prior-guidance, attention-based, and visual-language modeling approaches.
Given the strong correlation between pedestrian attributes and specific body components. Various methods, such as HPNet~\cite{2017pa100k} and DAHAR~\cite{2017pa100k, 2019dahar}, focused on localizing attribute-relevant regions via attention mechanisms. Variations in posture and viewpoint often challenge pedestrian attribute recognition. To address these challenges, some researchers~\cite{2021ssc,sarfraz2017deep} incorporated prior enhancement techniques or introduced supplementary neural networks to model these relationships effectively. Furthermore, pedestrian attributes are closely interconnected. Consequently, JLAC~\cite{2020JLAC} and PromptPAR~\cite{wang2023pedestrian} jointly model attribute context and image-attribute relationships.  While current methods recognize the importance of exploring contextual relationships in the PAR task, leveraging models like Transformers to capture attribute relationships within datasets often struggles to represent connections involving rare attributes.

\subsection{Benchmark Datasets for PAR}  
The most commonly used datasets of PAR are PETA~\cite{deng2014peta}, WIDER~\cite{li2016wider}, RAP~\cite{2016rapv1, 2019rapv2}, and PA100K~\cite{2017pa100k}. To enhance the ability to recognize pedestrian attributes at a long distance, Deng et al.~\cite{deng2014peta} introduced a new pedestrian attribute dataset named PETA, compiled from 10 small-scale pedestrian re-identification datasets, labeling over 60 attributes. Unlike PETA's identity-level annotation, the RAP dataset captures an indoor shopping mall and employs instance-level annotation for the pedestrian images. Both the PETA and RAPv1 datasets suffer from the issue of random segmentation, where individuals present in the training set also appear in the test set, resulting in information leakage. To solve this issue, Liu et al.~\cite{2017pa100k} proposed the largest pedestrian attribute recognition dataset in surveillance scenarios, PA100K, which contains 100,000 pedestrian images and 26 attributes. This dataset mitigates the information leakage problem by ensuring no overlap between pedestrians in the training and test sets. However, these datasets only contain simple scenes with limited background variation and lack significant style changes among pedestrians.

\subsection{Vision-Language Models}  
With the rapid development of the natural language processing field, many large language models (LLMs) such as Flan-T5~\cite{longpre2023flan}, and LLaMA~\cite{Touvron2023LLaMA} have emerged. Although notable foundational models like SAM~\cite{kirillov2023segment} have been introduced in the vision domain, the complexity of visual tasks has hindered the development of generalized multi-domain visual models. Some researchers have begun to view LLMs as world models, leveraging them as the cognitive core to enhance various multi-modal tasks. Recognizing the high cost of training a large multi-modal model from scratch, BLIP series~\cite{li2022blip, li2023blip2}, MiniGPT-4~\cite{zhu2024minigpt}, bridge existing pre-trained visual models and large language models. Although these models have significant improvements in the vision understanding and text generation field, there are many challenges, such as low-resolution image recognition, fine-grained image cation, and the hallucination of LLMs.

\begin{figure*}
\centering
\includegraphics[width=\textwidth]{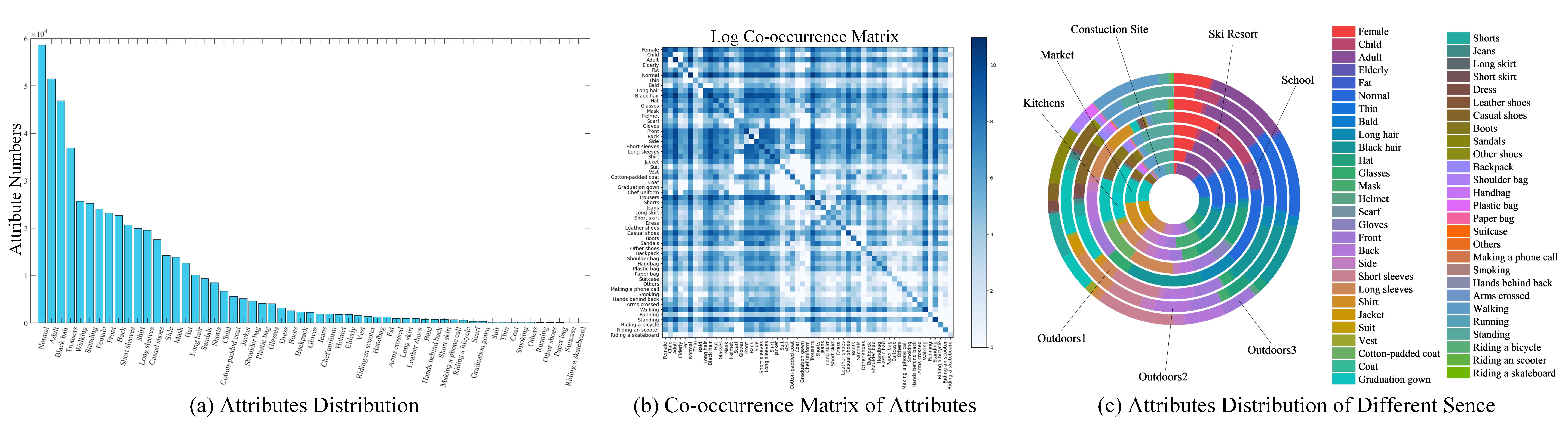}
\caption{
(a) Attributes Distribution: Bar graph showing the prevalence of individual attributes across the dataset; 
(b) Co-occurrence Matrix of Attributes: Logarithmic heatmap showing the co-occurrence frequency of attribute pairs; (c) Attributes Distribution in Different Scenes: Circular chart illustrating attribute distribution across eight different scenes.} 
\label{Dataset}
\end{figure*}

\section{MSP60K Benchmark Dataset} 

\subsection{Protocols}
To provide a robust platform for training and evaluating pedestrian attribute recognition (PAR) in real-world conditions, we adhere to these guidelines while constructing the MSP60K benchmark dataset: 
\emph{1). Large Scale:} We annotate 60,122 pedestrian images, each with 57 attributes, comprehensively analyzing pedestrian characteristics in various conditions.
\emph{2). Multiple Distances and Viewpoints:} Images are captured from different angles and distances using various cameras and handheld devices, covering the front, back, and side views. The resolution of pedestrian images in our dataset is from 30$\times$80 to 2005$\times$3008.
\emph{3). Complex and Varied Scenes:} Unlike existing datasets with uniform backgrounds, our dataset includes images from eight different environments with diverse backgrounds and attribute distributions, helping evaluate recognition methods in varied settings.
\emph{4). Rich Source of Pedestrian Identity:} We gather data on pedestrians from different scenarios, nationalities, and seasonal variations, enhancing the dataset with diverse styles and characteristics.
\emph{5). Simulated Complex Real-world Environments:} The dataset includes variations in lighting, motion blur, occlusions, and adverse weather conditions, simulating real-world challenges in pedestrian attribute recognition.

\begin{table}
\center
\small  
\caption{Comparison between our proposed MSP60K and existing PAR benchmark datasets.} \label{Publicdatasets} 
\begin{tabular}{l|c|c|c|c}
\hline \toprule [0.5 pt]
\multicolumn{1}{c|}{\multirow{1}{*}{\textbf{Dataset}}} & \multicolumn{1}{c|}{\textbf{Year}} & \multicolumn{1}{c|}{\textbf{Attributes}} & \multicolumn{1}{c|}{\textbf{Images}}  & \multicolumn{1}{c}{\multirow{1}{*}{\textbf{Scene Split}}}  \\
\hline    
PETA~\cite{deng2014peta} & 2014 & 61 & 19,000 & \XSolidBrush \\
WIDER~\cite{li2016wider} & 2016 & 14 & 57,524 & \XSolidBrush \\
RAPv1~\cite{2016rapv1} & 2016 & 69 & 41,585 & \XSolidBrush \\
PA100K~\cite{2017pa100k} & 2017 & 26 & 100,000 & \XSolidBrush \\
RAPv2~\cite{2019rapv2} & 2019 & 76 & 84,928 & \XSolidBrush \\
\hline 
Ours & 2024 & 57 & 60,015 & \Checkmark\\
\hline \toprule [0.5 pt] 
\end{tabular} 
\end{table}



\begin{table}[!htp]
\center
\scriptsize 
\caption{Attribute groups and details defined in our proposed MSP60K dataset.} 
\label{Attributes} 
\begin{tabular}{l|l}
\hline \toprule [0.5 pt]
\multicolumn{1}{c|}{\multirow{1}{*}{\textbf{Attribute Group}}} & \multicolumn{1}{c}{\textbf{Details}} \\
\hline     
Gender  & Female \\ \hline    
Age  & Child, Adult, Elderly \\ \hline
Body Size  & Fat, Normal, Thin \\ \hline
Viewpoint  & Front, Back, Side \\ \hline
Head  & \makecell[l]{Bald, Long Hair, Black Hair, Hat\\Glasses, Mask, Helmet, Scarf, Gloves} \\ \hline
Upper Body  & \makecell[l]{Short Sleeves, Long Sleeves, Shirt, Jacket, Suit, Vest\\Cotton 
Coat, Coat, Graduation Gown, Chef Uniform} \\ \hline
Lower Body  & \makecell[l]{Trousers, Shorts, Jeans, Long Skirt, Short Skirt, Dress} \\ \hline
Shoes  & \makecell[l]{Leather Shoes, Casual Shoes, Boots, Sandals, Other Shoes} \\ \hline
Bag  & \makecell[l]{Backpack, Shoulder Bag, Hand Bag\\ Plastic Bag, Paper Bag, Suitcase, Others} \\ \hline
Activity  & \makecell[l]{Calling, Smoking, Hands Back, Arms Crossed} \\ \hline
Posture  & \makecell[l]{Walking, Running, Standing, Bicycle, Scooter, Skateboard} \\
\hline \toprule [0.5 pt] 
\end{tabular} 
\end{table}

\subsection{Attribute Groups and Details}  
To effectively evaluate the performance of existing PAR methods in complex scenarios, each image in our dataset is labeled with 57 attributes, which are categorized into 11 groups: gender, age, body size, viewpoint, head, upper body, lower body, shoes, bag, body movement, and sports information. The complete list of the defined attributes can be found in Table~\ref{Attributes}.

\subsection{Statistical Analysis}  
As shown in Table~\ref{Publicdatasets}, MSP60K offers 8 distinct scenes and 57 attributes, providing richer annotations than datasets like PA100K (26 attributes) and WIDER (14 attributes). The dataset comprises 60,122 images of over 5,000 unique individuals. It includes varied environments such as markets, schools, kitchens, ski resorts, various outdoor and construction sites, offering a broader scope than other datasets.

In our benchmark dataset, we split the data using the random and cross-domain partitioning strategies: 

\noindent $\bullet$ \textbf{Random Partitioning}: 30,298 images for training, 6,002 for validation, and 23,822 for testing, ensuring a random distribution of scenes like other PAR benchmark datasets. 

\noindent $\bullet$ \textbf{Cross-domain Partitioning:} To validate domain generalization and zero-shot performance of PAR models, we divide our dataset based on scenarios, i.e., five scenarios (\textit{Construction Site, Market, Kitchens, School, Ski Resort}) with 34,128 images are used for training, while three scenarios (\textit{Outdoors1, Outdoors2, Outdoors3}) with 24,994 images are used for testing.

To assess the robustness of the model, we intentionally degrade 1/3 of the images in each subset by introducing variations such as changes in \textit{lighting}, \textit{random occlusions}, \textit{blurring}, and \textit{noise}. With its extensive size and diverse conditions, MSP60K offers a comprehensive platform for evaluating PAR methods.

\begin{figure*}
    \centering
    \includegraphics[width=1\linewidth]{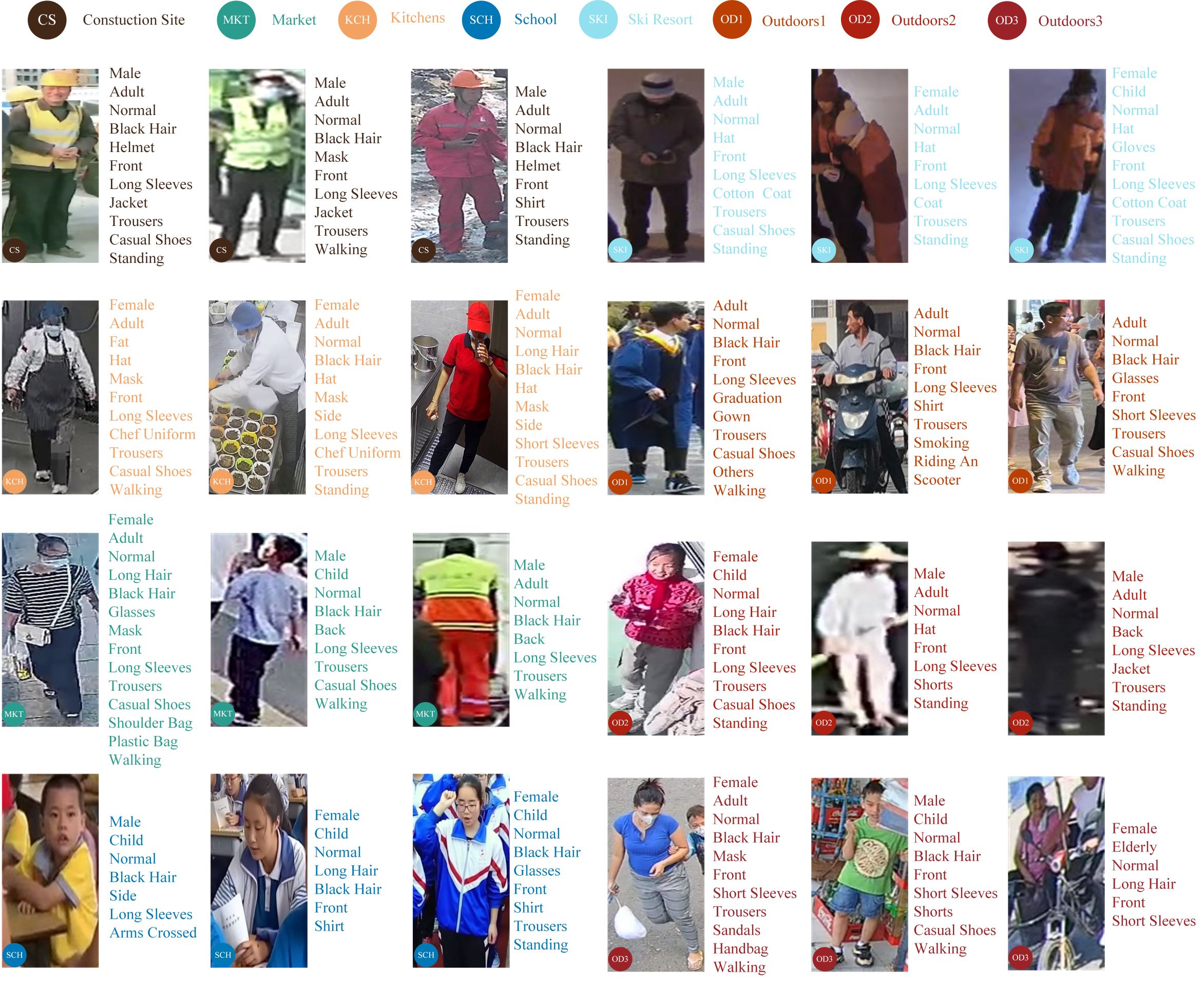}
    \caption{An illustration of representative samples in our newly proposed MSP60K PAR dataset.} 
    \label{fig:MSP60Ksamples}
\end{figure*}

\begin{figure*}
    \centering
    \includegraphics[width=1\linewidth]{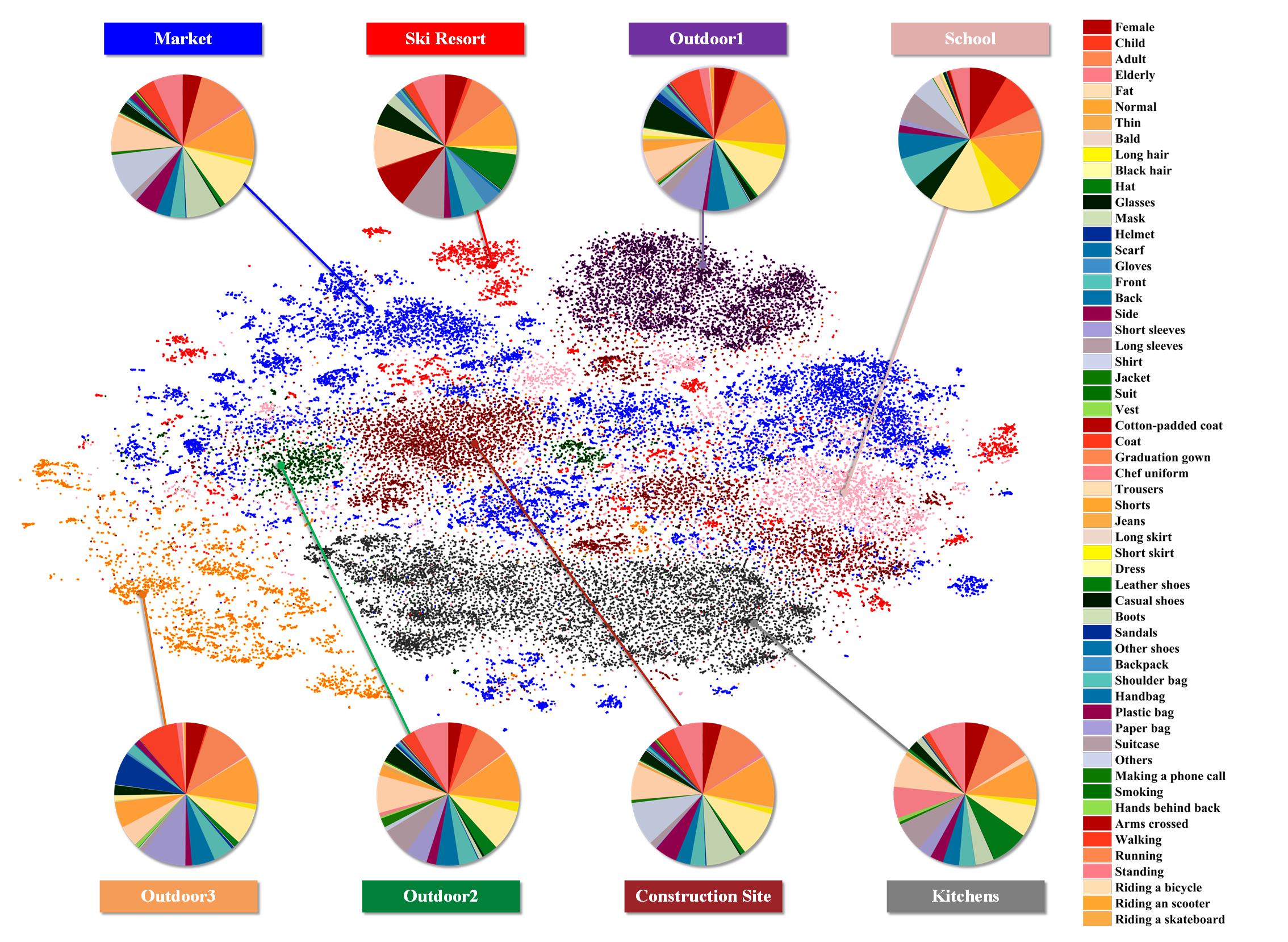}
    \caption{T-SNE visualization of scene samples in the MSP60K PAR dataset. Each colored cluster represents samples from different scenes, including ``Market," ``Ski Resort," ``Outdoor1," ``School," ``Outdoor3," ``Outdoor2," ``Construction Site", and ``Kitchens". For each scene, a pie chart is overlaid to illustrate the attribute distribution within that cluster. The legend on the right provides a detailed list of all attributes.} 
    \label{fig:MSP60Ktsne}
\end{figure*}

The dataset also exhibits a long-tail effect, similar to existing PAR datasets, depicted in Fig.~\ref{Dataset} (a),  and reflects real-world attribute distributions. Fig.~\ref{Dataset} (b) presents the co-occurrence matrix of pedestrian attributes, where each cell represents the frequency of two attributes appearing together. Darker areas indicate higher co-occurrence frequency. For example, \textit{Cotton-padded coat} and \textit{Long Sleeves} have a strong association, while attributes like \textit{Bald} and \textit{Long Hair/Black Hair} rarely co-occur. Fig.~\ref{Dataset} (c) displays the distribution of attributes across different scenarios, such as Construction Sites, Markets, Kitchens, and others, with attributes represented by different colors in a concentric circle plot. For instance, the \textbf{School} scenario has a higher number of \textit{Child} attributes, while the \textbf{Outdoors3} scenario shows a greater prevalence of \textit{Short Sleeves} and \textit{Sandals} attributes.

Fig.~\ref{fig:MSP60Ksamples} shows samples of every scene. It is evident that different scenes have various backgrounds, and the clothing styles have significant changes.

To visually demonstrate that each scene has a different distribution, we extract all image features using Resnet-50 and then visualize them using t-SNE, as shown in Figure~\ref{fig:MSP60Ktsne}. Clearly, the distributions of the different scenarios are roughly clustered in one area, and each scenario has a distinct attribute distribution. This visualization shows that the multi-scene segmentation of our dataset is meaningful.

\begin{figure*}
\centering
\includegraphics[width=0.98\textwidth]{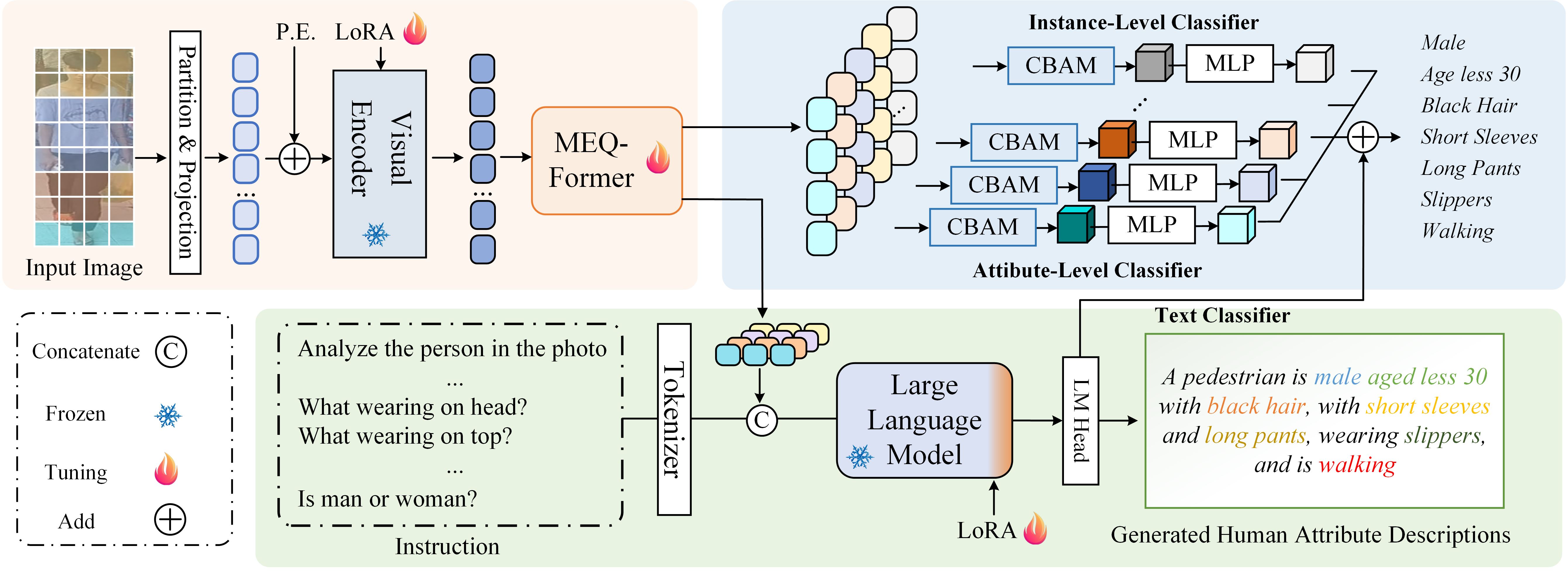}
\caption{An illustration of our proposed LLM-PAR framework illustrates how we use Multimodal Large Language Models (MLLMs) for deep semantic reasoning, combining images and descriptive text to provide more interpretable visual understanding. Through this framework, we can recognize pedestrian attributes and generate natural language descriptions, thereby offering more intuitive explanations. Our framework consists of three parts: visual feature extraction, language description generation, and language-enhanced classification.}
\label{framework}
\end{figure*}

\subsection{Benchmark Baselines} 
Our evaluation covers a variety of methods (17 total), including:   
\emph{1). CNN-based:} DeepMAR~\cite{deepmar}, RethinkPAR~\cite{2021Rethinking}, SSCNet~\cite{2021ssc}, SSPNet~\cite{SHEN2024110194}.
\emph{2). Transformer-based:} DFDT~\cite{ZHENG2023105708}, PARFormer~\cite{fan2023parformer}.
\emph{3). Mamba-based:} MambaPAR~\cite{wang2024SSMSurvey}, MaHDFT~\cite{wang2024empiricalmamba}.
\emph{4). Human-Centric Pre-Training Models for PAR:} PLIP~\cite{zuo2023plip}, HAP~\cite{yuan2024hap}.
\emph{5). Visual-Language Models for PAR:} VTB~\cite{cheng2022VTB}, Label2Label~\cite{li2022label2label}, PromptPAR~\cite{wang2023pedestrian}, SequencePAR~\cite{jin2023sequencepar}.

\section{Methodology} 
In this section, we introduce our proposed framework for pedestrian attribute recognition LLM-PAR. Our approach consists of three main parts: visual feature extraction, image caption generation, and the classification module. We first explain each of these three components. After that, we outline the training and inference process of our method.

\begin{figure}
\centering
\includegraphics[width=0.45\textwidth]{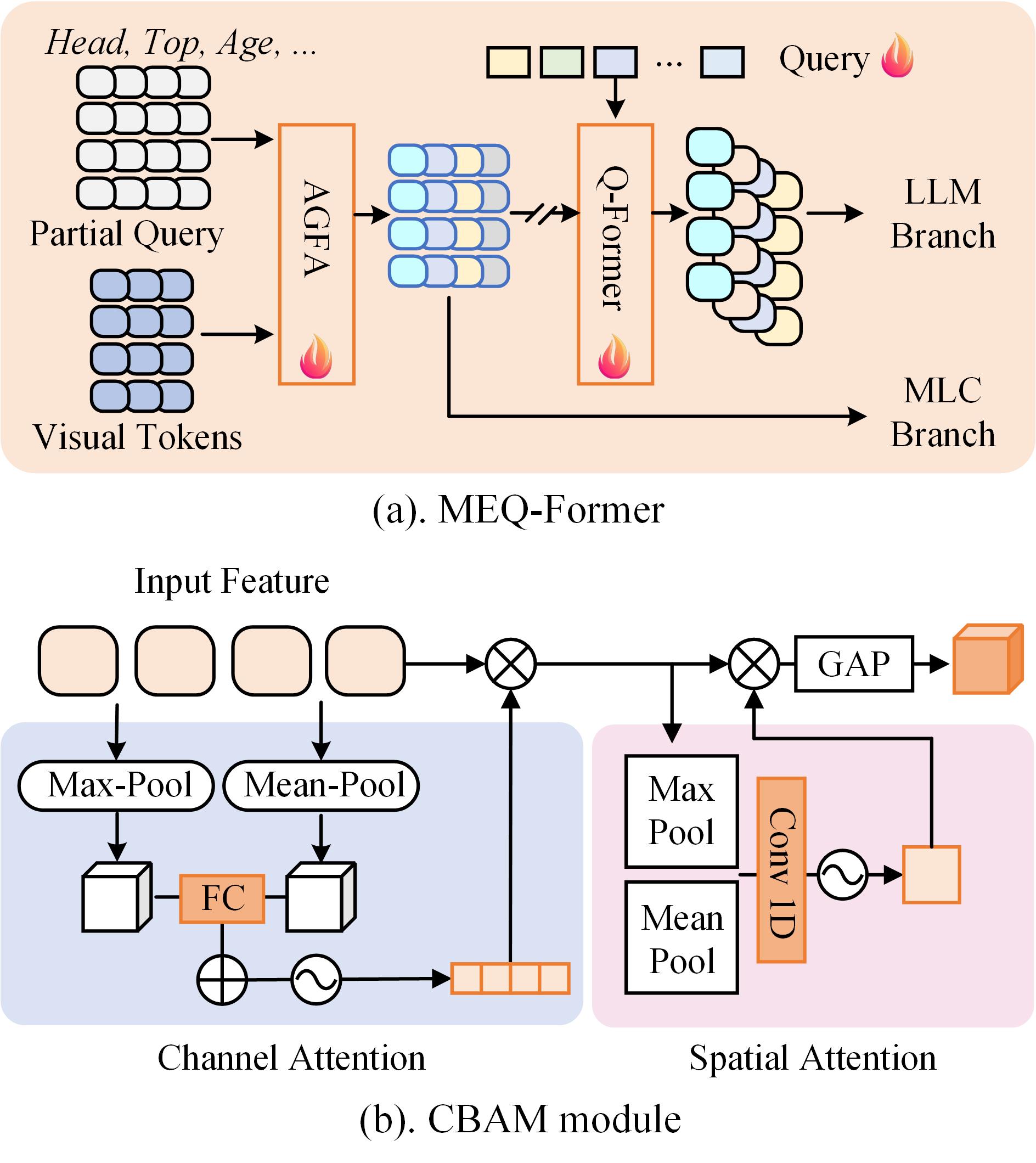}
\caption{The detailed architecture of (a). MEQ-Former and (b). CBAM module.}
\label{MEQFormerCBAM}
\end{figure}

\subsection{Overview}
This paper introduces a method for improving pedestrian attribute recognition (LLM-PAR) using multi-modal large language models (MLLMs) which describe the image in detail. As shown in Fig.~\ref{framework}, we leverage MLLMs to explore the contextual relationships between attributes, generating descriptions that assist attribute recognition. The approach consists of three main modules: 1) a multi-label classification branch, 2) a large language model branch, and 3) model aggregation. Specifically, we first extract the visual features of pedestrians using a visual encoder. Then, we design MEQ-Former to extract specific features for different attribute groups and translate to the latent space of MLLMs, improving the ability of MLLMs to capture fine details of pedestrians. The attribute group features are integrated into instruction embedding via a projection layer, the features feed into the large language model to generate pedestrian captions. Finally, the classification results from the visual features of each group are aggregated with the results from the language branch to produce the final classification results. The following sections will provide a detailed introduction to these modules.

\subsection{Multi-Label Classification Branch} 


Given an input pedestrian image $I \in \mathbb{R}^{H \times W \times 3}$, as shown in Fig.~\ref{framework}, we first partition it into patches and project them into visual tokens. The visual tokens are added with Position Embedding (P.E.) which encodes the spatial information. The output will be fed into a visual encoder (EVA-ViT-G~\cite{fang2023eva} is adopted for default) to extract the global visual representation $F_V$. In our implementation, we freeze the parameters of the pre-trained visual encoder and adopt LoRA~\cite{hu2022lora} to achieve efficient tuning. Then, a  newly designed Multi-Embedding Query Transformer (MEQ-Former) which extracts specific features from different attribute groups derived from primary visual features. Here, the attribute groups are obtained by categorizing the attributes into groups $A^j \mid j=\{0, 1, \dots, K\}$, based on their type, such as \textit{head}, \textit{upper body clothing}, \textit{actions}, where $K$ denotes the number of attribute groups.

As shown in Fig.~\ref{MEQFormerCBAM}, we create $K$ sets of Partial Query (PartQ) $Q_p \in \mathbb{R}^{K \times L \times D}$, where $L$ and $D$ are the number and dimension of the queries, respectively. These embeddings are fed into the Attributes Group Features Aggregate (AGFA) module to extract specific features $F_g=\{F_g^1, F_g^2, ..., F_g^K\}$ for different attribute groups. The AGFA module consists of stacked Feed-Forward Networks (FFN) and Cross-Attention (CrossAttn) layers. This process can be formulated as:  
\begin{equation}
\label{AGAF}
\small 
F_g = FFN(CrossAttn(Q=Q_p, K=F_V, V=F_V))
\end{equation} 
The $F_g$ is fed into the Q-Former $E_Q$, which serves as a bridge between the visual and language modalities, to generate text-related information $F_q^j$. Q-Former comprises stacked self-attention and cross-attention layers, and aggregates image information through cross-attention mechanisms. Then, we introduce the Convolutional Block Attention Modules (CBAM)~\cite{woo2018cbam} to capture fine-grained features for each attribute from the $F_g$ to produce attribute-specific predictions.

\subsection{Large Language Model Branch}



Although this multi-label classification framework can achieve decent accuracy, it still fails to consider the logical reasoning of large language models, which is evident in the image-text domain. Therefore, this paper attempts to use LLM as an auxiliary branch to enhance pedestrian attribute recognition. As shown in Fig.~\ref{framework}, we first build the instructions based on each attribute group $A^j$, i.e., 
\texttt{\textbf{Human}: Analyze the person's photo, and categorize it into attributes. \\ 
<Img><ImageHere\_\textit{Head}></Img> \\ 
What are wearing on their head? \\ 
<Img><ImageHere\_\textit{Topwear}></Img> \\ 
What are wearing on top? ...} \\ 
\texttt{\textbf{Assistant}:\underline{~~~~~~}.}    \\ 
Then, we adopt the \textit{Tokenizer}~\cite{zheng2023judging} to get the instruction embeddings $T_E = \{ T_E^1, T_E^2, \dots, T_E^{k+1} \}$ and concatenate them with visual features $F_q$ of the human image as the instruction features $F_I$. Note that, we embed the ground truth and concatenate it with $F_I$ as the initial input of the LLM during the training phase. The Vicuna-7B~\cite{zheng2023judging} and OPT-6.7B~\cite{liu2021opt} are exploited as the LLM and also tuned using LoRA in our experiments. Finally, we get the last hidden state from MLLM and the corresponding image captions through the Language Model Head.

\subsection{Model Aggregation for PAR}  
After being equipped with the LLM, our algorithmic framework can simultaneously output pedestrian attribute results and complete text passages to describe the attributes of a given pedestrian. To leverage the strengths of these two branches, we have designed an algorithm integration module to achieve enhanced prediction results. As shown in Fig.~\ref{framework}, we define two visual classifiers for attribute recognition, i.e., the attribute-level and instance-level classifiers. We also get the classifier for recognition using tokens from the large language model branch.

In our implementation, we exploit the following three strategies to fuse these three results as ours. Specifically, 
\textit{1). Attributes-Specific Aggregation (ASA)}: we adaptively weight and sum the attribute predictions of each classifier based on the weights learned from the training subset. 
\textit{2). Mean Pooling}: We directly take the average of the results from these three branches as the final model output. 
\textit{3). Max Pooling}: We take the maximum value of the logits from the three prediction branches as the final prediction result. 
Note that, we adopt the \textit{Mean Pooling} strategy as the default setting in our experiments if not otherwise specified.
More detailed results can be found in the sub-section~\ref{sec::AggStrategy} in our experiments.

\subsection{Loss Function} 
In the training phase, we adopt the widely used weighted cross-entropy loss (WCE Loss) $\mathcal{L}_{wce}(\cdot)$~\cite{deepmar} for attribute prediction branches, i.e.,
\begin{align}
\label{VisualLoss}
\mathcal{L}_{MLC} = \mathcal{L}_{wce}(\hat{y}, P_{attr}) + \mathcal{L}_{wce}(\hat{y}, P_{in}) 
\end{align}
We also adopt cross-entropy loss $\mathcal{L}_{ce}(\cdot)$ for the captioning generation in the LLM branch. 
\begin{align}
\label{captionLoss} 
\mathcal{L}_{LLM} = \mathcal{L}_{wce}(\hat{y},P_{llm}) + \mathcal{L}_{ce}(\hat{y}_{cap},P_{cap}) 
\end{align}
where $\hat{y}$ and $\hat{y}_{cap}$ denote the ground-truth labels and corresponding pedestrian attribute description, respectively. 
The $P_{cap}$ is the logits generated by the Large Language Model Head. 
More in detail, the $\mathcal{L}_{ce}(\cdot)$ and $\mathcal{L}_{wce}(\cdot)$ can be formulated as :
\begin{equation}
\label{CE2}
\mathcal{L}_{ce}(\cdot) =  -\frac{1}{M} \sum_{i=1}^M \text{CE}(y_{i}, p_{i})
\end{equation}
\begin{equation}
\label{WCE}
\mathcal{L}_{wce} =  -\frac{1}{M}\sum_{i=1}^Mw_i\text{CE}(y_{i}, p_{i})
\end{equation}
where $M$ is the number of attributes, $w_i$ is used to adjust the contribution for unbalanced categories, inversely related to the number of category positive samples. The \text{CE} term can be represented as: 
\begin{equation}
\label{CE1}
\text{CE}(y_{i}, p_{i}) =  y_{i}\log(p_{i})+(1-y_{i})\log(1-p_{i})
\end{equation}

\begin{table*}[!htb]
\center
\setlength{\tabcolsep}{5pt}
\small  
\caption{Comparison with public methods on our datasets. The \textcolor{red}{first} and \textcolor{blue}{second} are shown in \textcolor{red}{red} and \textcolor{blue}{blue}, respectively. Zero-shot refers to the use of MiniGPT4 for zero-shot inference to generate all dataset descriptions. It then utilizes BERT to extract text features, followed by training a fully connected layer for classification.} \label{Comparisononmspdatasets}
\begin{tabular}{l|c|c|ccccc|ccccc}
\hline \toprule [0.5 pt] 
\multirow{2}{*}{Methods} & \multirow{2}{*}{Publish} & \multirow{2}{*}{Code} & \multicolumn{5}{c|}{Random Split} & \multicolumn{5}{c}{Cross-domain Split} \\ \cline{4-13} 
 &
  \multicolumn{1}{c|}{} &
  \multicolumn{1}{c|}{} &
  \multicolumn{1}{c}{mA} &
  \multicolumn{1}{c}{Acc} & 
  \multicolumn{1}{c}{Prec} &
  \multicolumn{1}{c}{Recall} &
  \multicolumn{1}{c|}{F1} &
  \multicolumn{1}{c}{mA} &
  \multicolumn{1}{c}{Acc} &
  \multicolumn{1}{c}{Prec} &
  \multicolumn{1}{c}{Recall} &
  \multicolumn{1}{c}{F1} \\ \hline
\#01 DeepMAR~\cite{deepmar}  & ACPR15 & \href{https://github.com/dangweili/pedestrian-attribute-recognition-pytorch}{URL} & 70.46  & 72.83 & 84.71 & 81.46 & 83.06 & 54.84 & 44.97 & 63.38  & 58.81 & 61.01  \\
\#02 Strong Baseline~\cite{} & - & \href{https://github.com/aajinjin/Strong_Baseline_of_Pedestrian_Attribute_Recognition}{URL}  & 74.09 & 73.74 & 84.06 & 83.51 & 83.31 & 55.91 & 46.25  & 63.28 & 61.34 & 61.64 \\
\#03 RethinkingPAR~\cite{2021Rethinking} & arXiv20 & \href{https://github.com/valencebond/Rethinking_of_PAR}{URL} & 74.01  & 74.20 & 84.17 & 83.94 & 84.06 & 55.98 & 46.52 & 62.85 & 62.09 & 62.47  \\
\#04 SSCNet~\cite{2021ssc} & ICCV21 & \href{https://github.com/Geonu-Lee/Reimplementation_SSC}{URL} & 69.71  & 69.31  & 79.22  & 82.47  & 80.82 & 52.84 & 40.88  &  56.26 & 58.64 & 57.43 \\
\#05 VTB~\cite{cheng2022VTB}  & TCSVT22  & \href{https://github.com/cxh0519/VTB}{URL} & 76.09 & 75.36 & 83.56 & 86.46 & 84.56 & 58.59 & 49.81 & 65.11 & 66.11 & 65.00 \\
\#06 Label2Label~\cite{li2022label2label} & ECCV22 & \href{https://github.com/Li-Wanhua/Label2Label}{URL} & 73.61 & 72.66 & 81.79 & 84.32 & 82.56 & 56.38  & 45.81  & 59.67  & 64.20 & 61.19  \\
\#07 DFDT~\cite{ZHENG2023105708} & EAAI22 & \href{https://github.com/aihuazheng/DFDT}{URL} & 74.19 & 76.35 & \textcolor{red}{\textbf{85.03}} & 86.35 & 85.69 & 57.85 & 49.97 & 65.34 & 66.18 & 65.76 \\
\#08 Zhou et al.~\cite{Zhou2023Co} & IJCAI23 & \href{https://github.com/SDret/A-Solution-to-Co-occurrence-Bias-Attributes-Disentanglement-via-Mutual-Information-Minimization-for}{URL} & 73.07  & 68.76 & 78.38 & 82.10 & 80.20 & 54.26  & 41.91 & 56.23 & 60.11 & 58.11  \\
\#09 PARFormer~\cite{fan2023parformer} & TCSVT23 & \href{https://github.com/xwf199/PARFormer}{URL}  & 76.14 & 76.67 & 84.77 & 86.93 & 85.44 & 57.96 & 50.63 & 62.28 & 71.04 & 65.82 \\
\#10 SequencePAR~\cite{jin2023sequencepar} & arXiv23   &  \href{https://github.com/Event-AHU/OpenPAR}{URL} & 71.88 & 71.99 & 83.24 & 82.29 & 82.29 & 57.88 & 50.27 & 65.81 & 65.79 & 65.37 \\
\#11 VTB-PLIP ~\cite{zuo2023plip} & arXiv23  & \href{https://github.com/Zplusdragon/PLIP}{URL}  &73.90  &73.16  &82.01  &84.82  &82.93  &56.30  &46.77  &61.20  &64.47  &62.18  \\
\#12 Rethink-PLIP~\cite{zuo2023plip}  & arXiv23  & \href{https://github.com/Zplusdragon/PLIP}{URL}  &69.44  &68.90  &79.82  &81.15  &80.48  &57.18  &46.98  &63.57  &62.16  &62.86  \\
\rowcolor{blue!10}
\#13 PromptPAR~\cite{wang2023pedestrian}  & arXiv23  &  \href{https://github.com/Event-AHU/OpenPAR}{URL} & \textcolor{blue}{\textbf{78.81}} & \textcolor{blue}{\textbf{76.53}} & \textcolor{blue}{\textbf{84.40}} & \textcolor{blue}{\textbf{87.15}} & 85.35 & \textcolor{blue}{\textbf{63.24}} & \textcolor{blue}{\textbf{53.62}} & \textcolor{red}{\textbf{66.15}} & \textcolor{blue}{\textbf{71.84}} & \textcolor{blue}{\textbf{68.32}} \\
\#14 SSPNet \cite{zhou2024pedestrian} & PR24 &  \href{ https://github.com/guotengg/SSPNet}{URL} & 74.03  & 74.10  & 84.01  & 84.02  & 84.02  & 56.15  & 46.75  & 62.44  & 63.07  & 62.75  \\
\#15 HAP~\cite{yuan2024hap}  & NIPS24  & \href{https://github.com/junkunyuan/HAP}{URL} & 76.92 & 76.12 & 84.78 & 86.14 & \textcolor{blue}{\textbf{85.45}} & 58.70 & 50.59 & 65.60 & 66.91 & 66.25 \\
\#16 MambaPAR~\cite{wang2024SSMSurvey} & arXiv24 & \href{https://github.com/Event-AHU/OpenPAR/tree/main/MambaPAR_Empirical_Study}{URL} & 73.85  & 73.64 & 83.19 & 84.29 & 83.28 & 56.75 & 47.34 & 61.92 & 64.98 & 62.80 \\ 
\#17 MaHDFT~\cite{wang2024empiricalmamba} & arXiv24  & \href{https://github.com/Event-AHU/OpenPAR/tree/main/MambaPAR_Empirical_Study}{URL}  & 74.08  & 74.40  & 82.82  & 86.41  & 83.93 & 58.67 & 50.65 & 62.39 & 71.13 & 65.85\\
\hline 
Zero-shot & - & - & 56.93 & 52.97 & 72.26 & 64.69 & 67.46 & 52.19 & 39.26 & 60.12 & 52.09 & 55.15 \\
\rowcolor{red!10}
Ours &- & - & \textcolor{red}{\textbf{80.13}} & \textcolor{red}{\textbf{78.71}} & 84.39 & \textcolor{red}{\textbf{90.52}} & \textcolor{red}{\textbf{86.94}} & \textcolor{red}{\textbf{66.29}} & \textcolor{red}{\textbf{58.11}} & \textcolor{blue}{\textbf{65.68}} & \textcolor{red}{\textbf{81.21}} & \textcolor{red}{\textbf{72.05}} \\ 
\hline \toprule [0.5 pt] 
\end{tabular}
\end{table*}

\section{Experiments}

\subsection{Datasets and Evaluation Metric}  

In this study, we conduct a comprehensive benchmark of 17 pedestrian attribute recognition methods, representing the most important models in the field of pedestrian attribute recognition. Furthermore, the performance of our methods is compared with existing state-of-the-art (SOTA) PAR methods in our benchmark and in three publicly available datasets: \textbf{PETA}~\cite{deng2014peta}, \textbf{PA100K}~\cite{2017pa100k} and \textbf{RAPv1}~\cite{2016rapv1}. Five widely used evaluation metrics are employed for evaluating the performance, including: 
\textbf{mean Accuracy}~(mA), 
\textbf{Accuracy}~(Acc), 
\textbf{Precision}~(Prec), 
\textbf{Recall} and 
\textbf{F1-score}~(F1).
More details about these evaluation metrics can be found in our supplementary materials.

\subsection{Implementation Details}

In the training phase, we use ground truth to expand the attributes as appropriate sentences by the template, creating the $<$instruction, answer$>$ set to fine-tune the LLM. Additionally, we utilize the ground-truth sentences mask strategy to prevent information leakage during the training stage, which helps in effectively learning the LLM classification head. For inference, we auto-regressive generate the sentence from the instruction with the image feature, and use the last step hidden state that predicts the result of the language branch.

We utilize EVA-ViT-G~\cite{fang2023eva} as the visual backbone, and its last three layers are used to initial the AGFA module. The Q-Former adopts BERT~\cite{kenton2019bert} with several cross-attention layers added to interact with visual features. and we default utilize Vicuna-7B~\cite{zheng2023judging} as the large language model. All backbones are initialized according to the MiniGPT-4~\cite{zhu2024minigpt} settings and weights. We adopt LoRA~\cite{hu2022lora} to fine-tune the visual backbone and the last 3 layers of LLMs. The LoRA is only injected in the projection of $Q$ and $V$ in the attention layer, with the low-rank dimension $r$ set as 32. We train the models for 60 epochs using the AdamW optimizer, with a learning rate of 0.00002 and a weight decay of 0.0001. The training is conducted on a server with NVIDIA A800-SXM4-80GB with a batch size of 4. More details can be found in our source code.

\subsection{Comparison on Public PAR Benchmarks}

\noindent $\bullet$ \textbf{Result on MSP60K Dataset.} 
We collect and analyze public PAR methods from 2015 to 2024 on the MSP60K dataset as shown in Table~\ref{Comparisononmspdatasets}, methods like HAP~\cite{yuan2024hap}, RethinkingPAR~\cite{2021Rethinking}, and PARformer~\cite{fan2023parformer}, which perform well in the random split but experience significant drops in performance in the cross-domain split. For instance, mA, Acc, and F1 of HAP scores drop by 18.22, 25.53, and 19.20, respectively. Some methods show smaller declines in the cross-domain split, with PromptPAR~\cite{wang2023pedestrian} achieving state-of-the-art results, though still with notable decreases. We also test MiniGPT-4~\cite{zhu2024minigpt} in a zero-shot setup on our dataset, with significant drops observed in the cross-domain split. After optimizations, LLM-PAR achieves 80.13, 78.71, 84.39, 90.52, and 86.94 in the random split, and 66.29, 58.11, 65.28, 81.21, and 72.05 in the cross-domain split, which achieves the best results on nearly all metrics. The experiments on the MSP60K dataset fully validate the effectiveness of our proposed LLM-PAR for attribute recognition.


\begin{table*}[!htb]
\center
\caption{Comparison with SOTA methods on PETA, PA100K and RAPv1 datasets. The \textcolor{red}{first} and \textcolor{blue}{second} are shown in \textcolor{red}{red} and \textcolor{blue}{blue}, respectively.} \label{Comparisononpublicdatasets} 
\resizebox{1\textwidth}{!}{
\begin{tabular}{l|c|ccccc|ccccc|ccccc}
\hline \toprule [0.5 pt] 
\multicolumn{1}{c|}{\multirow{2}{*}{Methods}} & \multicolumn{1}{c|}{\multirow{2}{*}{Publish}}   & \multicolumn{5}{c|}{PETA} & \multicolumn{5}{c|}{PA100K} & \multicolumn{5}{c}{RAPv1}\\ \cline{3-17} 
\multicolumn{1}{c|}{} &
\multicolumn{1}{c|}{} &
  \multicolumn{1}{c}{mA} &
  \multicolumn{1}{c}{Acc} & 
  \multicolumn{1}{c}{Prec} &
  \multicolumn{1}{c}{Recall} &
  \multicolumn{1}{c|}{F1} &
  \multicolumn{1}{c}{mA} &
  \multicolumn{1}{c}{Acc} &
  \multicolumn{1}{c}{Prec} &
  \multicolumn{1}{c}{Recall} &
  \multicolumn{1}{c|}{F1}  &
  \multicolumn{1}{c}{mA} &
  \multicolumn{1}{c}{Acc} &
  \multicolumn{1}{c}{Prec} &
  \multicolumn{1}{c}{Recall} &
  \multicolumn{1}{c}{F1}  \\ \hline
SSCsoft~\cite{2021ssc} & ICCV21 & 86.52 & 78.95 & 86.02 & 87.12 & 86.99 & 81.87 & 78.89 & 85.98 & 89.10 & 86.87 & 82.77 & 68.37 & 75.05 & 87.49 & 80.43\\ 				
IAA~\cite{2022iaacaps} & PR22 & 85.27 & 78.04 & 86.08 & 85.80 & 85.64 & 81.94 & 80.31 & 88.36 & 88.01 & 87.80 & 81.72 & 68.47 & 79.56 & 82.06 & 80.37\\
MCFL~\cite{Chen2022MCFL} &  NCA22 & 86.83 & 78.89 & 84.57 & 88.84 & 86.65 & 81.53 & 77.80 & 85.11 & 88.20 & 86.62 & 84.04 & 67.28 & 73.44 &87.75 & 79.96\\
DRFormer~\cite{2022drformer} &  NC22 & 89.96 &81.30 & 85.68 &91.08 &88.30 & 82.47 & 80.27 & 87.60 & 88.49 & 88.04 & 81.81 &70.60 & 80.12 & 82.77 & 81.42\\
VAC~\cite{guo2022visual} & IJCV22  & - & - & - & - & - & 82.19 & 80.66 & 88.72 & 88.10 & 88.41 & 81.30 & 70.12 &81.56 & 81.51 &81.54\\
DAFL~\cite{jia2022learning} & AAAI22 & 87.07 & 78.88 & 85.78 & 87.03 & 86.40 & 83.54 & 80.13 & 87.01 & 89.19 & 88.09 & 83.72 & 68.18 & 77.41 & 83.39 & 80.29  \\
CGCN~\cite{Fan2022CGCN} & TMM22  & 87.08 & 79.30 & 83.97 & 89.38 & 86.59 & - & - & - & - & - & 84.70 & 54.40 & 60.03 & 83.68 & 70.49 \\ 
CAS~\cite{yang2021cascaded} &  IJCV22 & 86.40 & 79.93 & 87.03 & 87.33 & 87.18 & 82.86 & 79.64 & 86.81 & 87.79 & 85.18 & 84.18 & 68.59 & 77.56 & 83.81 & 80.56\\  
VTB~\cite{cheng2022VTB} & TCSVT22  & 85.31 & 79.60 & 86.76 & 87.17 & 86.71 & 83.72 & 80.89 & 87.88 & 89.30 & 88.21 & 82.67 & 69.44 & 78.28 & 84.39 & 80.84 \\
\rowcolor{blue!10}
PromptPAR~\cite{wang2023pedestrian} & arXiv23 &88.76 &82.84 & \textcolor{red}{\textbf{89.04}} &89.74 &89.18 &87.47 &\textcolor{blue}{\textbf{83.78}} &\textcolor{red}{\textbf{89.27}} & \textcolor{blue}{\textbf{91.70}} &\textcolor{blue}{\textbf{90.15}}  &85.45 &\textcolor{blue}{\textbf{71.61}} & 79.64 & 86.05 &82.38  \\
PARformer~\cite{fan2023parformer} & TCSVT23 &89.32 &82.86 &88.06 & \textcolor{blue}{\textbf{91.98}} &89.06 &84.46 &81.13 &88.09 &91.67 &88.52 & 84.43 &69.94 & 79.63 & \textcolor{blue}{\textbf{88.19}} &81.35 \\
\rowcolor{blue!10}
OAGCN~\cite{lu2023oagcn} & TMM23   &\textcolor{blue}{\textbf{89.91}} &\textcolor{blue}{\textbf{82.95}} &88.26 &89.10 &88.68 &83.74 &80.38 &84.55 &90.42 &87.39  &\textcolor{red}{\textbf{87.83}} &69.32 &78.32 &87.29 & \textcolor{blue}{\textbf{82.56}} \\
\rowcolor{blue!10}
SSPNet~\cite{SHEN2024110194} & PR24 & 88.73 & 82.80 & \textcolor{blue}{\textbf{88.48}} & 90.55 & \textcolor{blue}{\textbf{89.50}} & 83.58 & 80.63 & 87.79 & 89.32 & 88.55 & 83.24 & 70.21 & \textcolor{red}{\textbf{80.14}} & 82.90 & 81.50 \\
SOFA~\cite{wu2024selective} & AAAI24  &87.10 &81.10 &87.80 &88.40 &87.80 & 83.40 &81.10 &\textcolor{blue}{\textbf{88.40}} &89.00 &88.30 &83.40 &70.00 &\textcolor{blue}{\textbf{80.00}} &83.00 &81.20   \\
FRDL~\cite{zhou2024pedestrian} & ICML24  &88.59 &- &- &- &89.03 &\textcolor{blue}{\textbf{89.44}} &- &- &- &88.05 &87.72 &- &- &- &79.16 \\
\hline
Zero-shot & - & 61.32 & 50.75 & 68.57 & 64.00 & 65.52 & 65.26 & 56.99 & 79.21 & 65.20 & 70.75 & 65.46 & 50.90 & 64.48 & 65.20 & 66.06 \\
\rowcolor{red!10}
Ours & - & \textcolor{red}{\textbf{92.25}} & \textcolor{red}{\textbf{84.59}} & 88.41 & \textcolor{red}{\textbf{92.94}} & \textcolor{red}{\textbf{90.39}} & \textcolor{red}{\textbf{91.09}} & \textcolor{red}{\textbf{84.12}} & 87.73 & \textcolor{red}{\textbf{94.09}} & \textcolor{red}{\textbf{90.41}}  & \textcolor{blue}{\textbf{87.80}} & \textcolor{red}{\textbf{71.86}} & 78.36 & \textcolor{blue}{\textbf{88.20}} & \textcolor{red}{\textbf{82.64}} \\
\hline \toprule [0.5 pt] 
\end{tabular}}
\end{table*}

\noindent $\bullet$ \textbf{Result on PETA~\cite{deng2014peta} Dataset.} 
As shown in Table~\ref{Comparisononpublicdatasets}, our method significantly outperforms previous methods. Compared to the previous best method SSPNet~\cite{SHEN2024110194} with prior guidance, we observe improvements of 3.52, 1.79, and 0.89 in mA, Acc, and F1, respectively. This illustrates the effectiveness of MLLMs without fine-tuned design in PAR. In contrast to PromptPAR with visual-language modeling by Transformer~\cite{vaswani2017Former} and CLIP~\cite{radford2021CLIP}, we also improve in 3.49, 1.75, and 1.21. 


\noindent $\bullet$ \textbf{Result on PA100K~\cite{2017pa100k} Dataset.} 
As shown in Table~\ref{Comparisononpublicdatasets}, our method also achieves optimal results on larger datasets, exceeding 1.65 and 2.36 on the mA and F1 metrics, respectively, compared to recent methods such as FRDL~\cite{zhou2024pedestrian}, without employing any resampling strategy. Compared to Transformer-based methods like PARformer~\cite{wang2023pedestrian}, our method shows a significant advantage, with results of 91.09, 84.12, and 90.41 on mA, Accuracy, and F1, respectively. Additionally, the progress is substantial compared to zero-shot MiniGPT~\cite{zhu2024minigpt}.

\noindent $\bullet$ \textbf{Result on RAPv1~\cite{2016rapv1} Dataset} 
Our framework obtains the SOTA performance compared with existing methods. Compared with the SOTA method OAGCN~\cite{lu2023oagcn} with using additional information of viewpoint, our method gets 87.80, 71.86, 78.36, 88.20, and 82.64, while the OAGCN gets 87.83, 69.32, 78.32, 87.29, and 82.56, and exceeds 4.40, 1.86, and 1.44 contrast to the SOFA~\cite{wu2024selective}. 

Based on the experiments conducted on the four datasets, it is clear that LLM-PAR delivers impressive results by combining visual classification and LLM modeling within the LLM-augment framework. Furthermore, the AGFA module extracts attribute group-specific features to capture detailed information and integrate them with Q-former into MEQ-Former, thereby enhancing the pedestrian caption details of LLMs.

\begin{table*}[htp]
\center
\small      
\caption{Component Analysis on the PETA Dataset. mA, Acc, and F1 results are reported. AGFA denote the Attributes Group Features Aggregation.} 
\label{ablation} 
\begin{tabular}{l|cccccc|ccc} 		
\hline \toprule [0.5 pt] 
\multicolumn{1}{c|}{\multirow{2}{*}{\#}} & \multicolumn{1}{c}{\multirow{2}{*}{CLS-Attr}} & \multicolumn{1}{c}{\multirow{2}{*}{FT Q-Former}} & \multicolumn{1}{c}{\multirow{2}{*}{LoRA}}   & \multicolumn{1}{c}{\multirow{2}{*}{CLS-LLM}} & \multicolumn{1}{c}{\multirow{2}{*}{AGFA}}   & \multicolumn{1}{c|}{\multirow{2}{*}{CLS-IN}}  & \multicolumn{3}{c}{PETA Dataset} \\ \cline{8-10} 
\multicolumn{1}{c|}{} &
\multicolumn{1}{c}{} &
\multicolumn{1}{c}{} &
\multicolumn{1}{c}{} &
\multicolumn{1}{c}{} &
\multicolumn{1}{c}{} &
\multicolumn{1}{c|}{} &
\multicolumn{1}{c}{mA} &
\multicolumn{1}{c}{Acc} &
\multicolumn{1}{c}{F1}
\\ 
\hline 
1   &\checkmark &           &           &           &           &           & 71.54 & 58.24 & 71.96 \\
2   &\checkmark &\checkmark &           &           &           &           & 82.89 & 72.32 & 81.89 \\ \hline 
3   &\checkmark &\checkmark &\checkmark &           &           &           & 90.14 & 83.25 & 89.38 \\
4   &\checkmark &\checkmark &\checkmark &\checkmark &           &           & 90.89 & 83.64 & 89.60 \\
5   &\checkmark &\checkmark &\checkmark &\checkmark &\checkmark &           & 91.78 & 84.47 & 90.27 \\
6   &\checkmark &\checkmark &\checkmark &\checkmark &\checkmark &\checkmark & 92.25 & 84.59 & 90.39 \\ 
\hline

\hline \toprule [0.5 pt]
\end{tabular}
\end{table*}

\subsection{Component Analysis}  
We conduct ablation experiments to analyze the contributions of different components in our method, including the visual backbone, AGFA module, LLM branch, and CLS-IN module. The visual backbone analysis reveals that the EVA-CLIP~\cite{fang2023eva} and Q-Former~\cite{li2023blip2} alone achieve mA, Acc, and F1 scores of 71.54, 58.24, and 71.96, respectively. Fine-tuning with LoRA~\cite{hu2022lora} improves these scores to 90.14, 83.25, and 89.38. The LLM branch alone achieves scores of 90.89, 83.64, and 89.60, which further improve to 92.20, 83.76, and 89.70 when combined with the AGFA module, demonstrating the effectiveness of LLMs in enhancing attribute recognition and their complementarity with the visual branch. The efficacy of the AGFA module is confirmed with scores of 92.20, 83.76, and 89.70, highlighting its role in improving feature aggregation and model recognition capabilities. Lastly, the CLS-IN module improves the mA, Acc, and F1 scores by 0.22, 0.12, and 0.13, respectively, indicating its contribution to enhancing the recognition of tail categories and supplementing other categories through shared feature learning.

\begin{figure*}
\centering
\includegraphics[width=\textwidth]{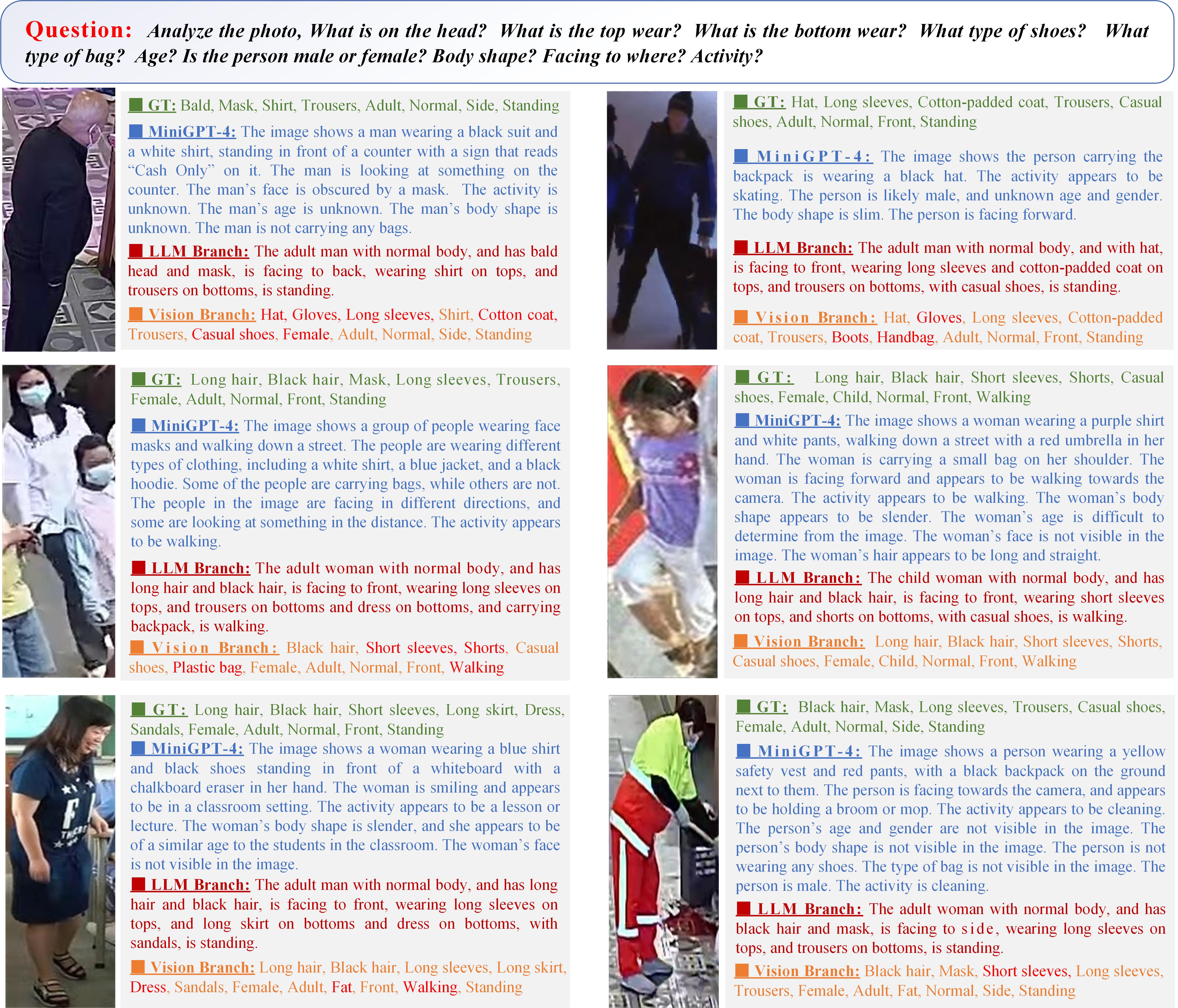}
\caption{Comparison of the caption and recognition results of our LLM-PAR and MiniGPT-4.}
\label{Caption}
\end{figure*}

\subsection{Ablation Study} 
In this section, we conduct detailed analysis experiments on the main module of LLM-PAR. This includes Ground-Truth Mask Strategies, the Number of AGFA Layers, the Length of PartQ, the Aggregation Strategy of Three Branches, and Different MLLMs in the PETA~\cite{deng2014peta} dataset.


\begin{table}[htb]
\center
\small  
\setlength{\tabcolsep}{3pt}
\caption{Comparing different ground truth replacement strategies. } \label{replacement} 
\begin{tabular}{l|cc|cc}
\hline \toprule [0.5 pt]
\multicolumn{1}{c|}{\multirow{2}{*}{Replacement}} &  \multicolumn{2}{c|}{CLS-Mean} &  \multicolumn{2}{c}{CLS-LLM} \\ \cline{2-5} 
\multicolumn{1}{c|}{} &
\multicolumn{1}{c}{mA} &
\multicolumn{1}{c|}{F1} &
\multicolumn{1}{c}{mA} &
\multicolumn{1}{c}{F1} \\ \hline
\#1 Ground Truth & 91.53 & 86.11 & 77.03 & 70.91 \\
\#2 25\% Mask(Padding) & 92.15  & 89.44 &  86.90 & 87.35 \\
\#3 50\% Mask(Padding) & 92.33 & 89.21 & 88.20 & 88.07 \\
\#4 75\% Mask(Padding) & 92.25 & 89.23 & 86.64 & 85.49  \\
\#5 100\% Mask(Padding) & 91.70 & 89.64 & 64.59 & 65.43 \\
\#6 Random Sentence & 92.25 & 90.39 & 88.84 & 89.22 \\
\hline \toprule [0.5 pt] 
\end{tabular}
\end{table}

\noindent $\bullet$ \textbf{Analysis on the Ground-Truth Mask Strategies.}
During the training phase, we observe that using ground truth directly for fine-tuning the language model leads to poor generalization due to information leakage. To improve this, we introduce a ground truth masking strategy. We compare various masking approaches to using ground truth directly (see Table~\ref{replacement}). Direct use of ground truth results in poor performance in the language branch during testing. Random masking of sentence is also ineffective, with high masking rates hindering meaningful sentence generation. The best results are obtained with a 50\% masking rate, improving mA and F1 scores by 0.80 and 3.10, respectively. Replacing ground truth with random sentences from the training set yielded the best performance. This strategy likely increases training difficulty, encouraging the model to utilize attribute context and visual information for better error correction.

\begin{table}[t]
\centering
\small
\setlength{\tabcolsep}{3pt}
\caption{Performance Comparison for AGFA Across Different Layers and Query Numbers}
\label{ComparingAGFA}
\resizebox{0.45\textwidth}{!}{%
\begin{tabular}{c|ccccc|ccc}
\hline
   \multirow{2}{*}{AGFA} & \multicolumn{5}{c|}{Layers} & \multicolumn{3}{c}{Querys} \\ 
\cline{2-9}
    & 1 & 3 & 6 & 9 & 12 & 64 & 128 & 256 \\
\hline
mA  & 91.97 & 92.25 & 92.57 & 92.68 & 92.77 & 92.01 & 92.25 & 92.20 \\ 
F1  & 89.82 & 90.39 & 90.61 & 90.69 & 90.53 & 88.28 & 90.39 & 90.02 \\ 
\hline
\end{tabular}
}
\end{table}

\noindent $\bullet$ \textbf{Analysis on the Number of AGFA Layers.}
As shown in Table~\ref{ComparingAGFA}, we introduce the AGFA module for extracting pedestrian attribute group features in this study. We analyze the impact of AGFA modules with 1, 3, 6, 9, and 12 layers on recognition performance. Our analysis reveals that increasing the number of AGFA layers improved recognition performance. However, considering computational efficiency, we opt for a 3-layer AGFA module to balance computational burden and performance.

\noindent $\bullet$ \textbf{Analysis on the Length of PartQ.}
As shown in Table~\ref{ComparingAGFA}, we examine the effect of the number of attribute group queries in the AGFA module on performance. Our findings show that using 128 queries obtains the best performance, with performance deteriorating with more than 256 queries and a significant decline observed with 64 queries.

\begin{table}[!htb]
\centering
\small
\setlength{\tabcolsep}{3pt}
\caption{Comparison of different aggregation strategies of logits.} 
\label{aggregation}
\begin{tabular}{c|c|c|c}
\hline \toprule [0.5 pt]
\multicolumn{1}{c|}{Metric} & \multicolumn{1}{c|}{ASA} & \multicolumn{1}{c|}{Mean Pooling} & \multicolumn{1}{c}{Max Pooling} \\ \hline
mA  & 91.53 & 92.25 & 92.46 \\ 
F1  & 90.17 & 90.39 & 88.95 \\ 
\hline \toprule [0.5 pt]
\end{tabular}
\end{table}

\begin{figure*}
\centering
\includegraphics[width=\textwidth]{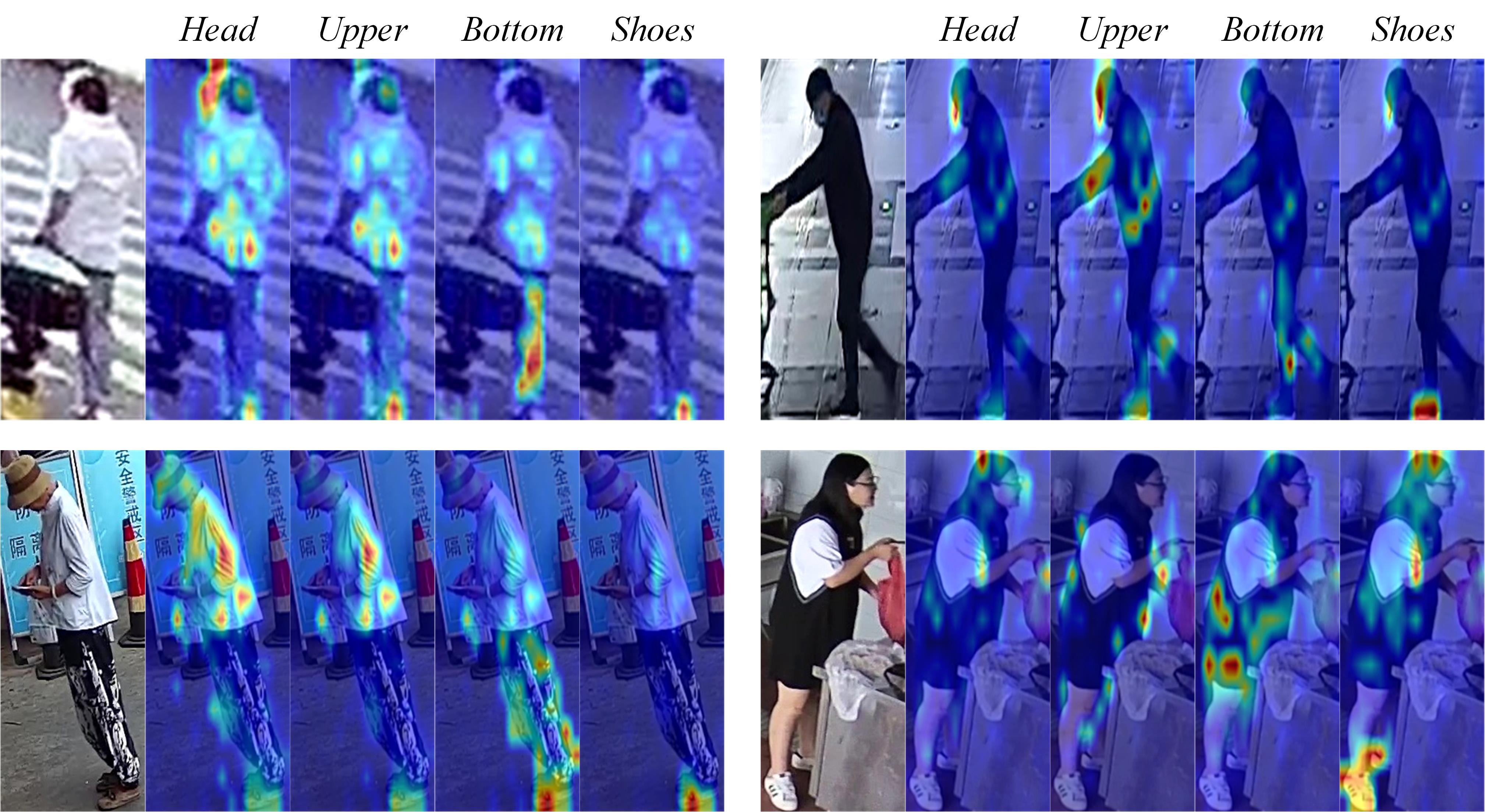}
\caption{The feature map of AGFA}\label{featuremap}
\end{figure*}

\noindent $\bullet$ \textbf{Analysis on the Aggregation Strategy of Threes Branches.} \label{sec::AggStrategy}
To improve the aggregation of results from three branches, we design and evaluate some aggregation strategies, including mean pooling and max pooling, and the performance of each strategy is reported in Table~\ref{aggregation}. Mean pooling achieves 92.20 and 90.02 in mA and F1 scores, respectively, while max pooling achieves 92.46 and 88.95. We find that mean pooling mitigates the influence of abnormal values on the final result. Additionally, we explore and design the attribute-specific aggregation (ASA) using the training datasets to obtain attribute-level fusion weights for the three branches, resulting in 91.53 and 90.17.

\begin{table}[!htb]
\centering
\small
\caption{Comparing of using different LLMs} \label{llms}
\begin{tabular}{l|c|c}
\hline \toprule [0.5 pt]
\multicolumn{1}{c|}{LLMs} & \multicolumn{1}{c|}{Vicuna-7B}  & \multicolumn{1}{c}{OPT-6.7B} \\ \hline
mA  & 92.25 & 92.12 \\ 
F1  & 90.39 & 89.39 \\ 
\hline \toprule [0.5 pt]
\end{tabular}
\end{table}

\noindent $\bullet$ \textbf{Analysis on the Different MLLMs.}
As shown in Tab.~\ref{llms}, we incorporate different MLLMs, such as Vicuna-7B~\cite{zheng2023judging} and OPT-6.7B~\cite{liu2021opt}, into our frameworks. The performance of LLM-PAR experiences both degradation and enhancement when we replace the MLLM as OPT, LLM-PAR achieved 92.12 and 89.39.


\subsection{Visualization} 



\noindent $\bullet$ \textbf{Recognition Results.}
In Fig.~\ref{Caption}, we present the findings and descriptions of LLM-PAR. Our baseline model, MiniGPT-4~\cite{zhu2024minigpt}, can broadly describe pedestrians, including gender and accessories. Still, it can cause severe hallucinations, such as the first image: \textit{standing in front of a counter with a sign that reads “Cash Only” on it} is not in the picture and the wrong prediction of gender in the last image: \textit{The person is male}. Conversely, our LLM-PAR is capable of accurately recognizing specific attributes of pedestrians.

\noindent $\bullet$ \textbf{Feature Map.}
As shown in Fig.~\ref{featuremap}, we display the feature map between PartQ and visual features in the AGFA module. This visualization demonstrates that PartQ accurately focuses on the pedestrian region, such as the \textit{Bottom} and \textit{Shoes} Query is obviously concerned about the trousers and shoes part of the pedestrians.


\section{Conclusion}  
This paper addresses the limitations of existing pedestrian attribute recognition (PAR) datasets by introducing MSP60K, a new large-scale, cross-domain dataset with 60,122 images and 57 attribute annotations across eight scenarios. By incorporating synthetic degradation, we further bridge the gap between the dataset and real-world challenging conditions. Our comprehensive evaluation of 17 representative PAR models under both random and cross-domain split protocols establishes a more rigorous benchmark. Moreover, we propose the LLM-PAR framework, which leverages a pre-trained vision Transformer backbone, a multi-embedding query Transformer for partial-aware feature learning, and is enhanced by a Large Language Model for ensemble learning and visual feature augmentation. The experimental results across multiple PAR benchmark datasets demonstrate the effectiveness of our proposed framework. Both the MSP60K dataset and the source code will be released to the public upon acceptance, contributing to future advancements in human-centered research and PAR technology. 

In our future work, we plan to further expand the scale of the dataset to conduct more extensive and thorough experimental validations. Moreover, the training and inference of the model still require substantial computational resources. In the future, we will design lightweight models to achieve a better balance between accuracy and performance. 



{
    \small
    \bibliographystyle{ieeenat_fullname}
    \bibliography{reference}
}

\end{document}